\documentclass[a4paper,conference]{IEEEtran}
\ifCLASSINFOpdf
\else
\fi

\usepackage{times}
\usepackage{helvet}
\usepackage{courier}
\usepackage{bm}
\usepackage{amsmath}
\usepackage{amssymb}
\usepackage{graphicx}
\usepackage{xcolor}
\usepackage{tabularx,ragged2e,booktabs,caption}

\begin{document}
%

\title{Nonlinear Metric Learning through Geodesic Interpolation within Lie Groups}

\author{ \IEEEauthorblockN{Zhewei Wang}
  \IEEEauthorblockA{School of EECS \\
    Ohio University\\
    Athens, OH 45701} \and \IEEEauthorblockN{Bibo Shi}
  \IEEEauthorblockA{Department of Radiology\\
    Duke University \\
    Durham, NC 27710} \and \IEEEauthorblockN{Charles D. Smith}
  \IEEEauthorblockA{Department of Neurology\\
    University of Kentucky\\
    Lexington, KY 40508} \and \IEEEauthorblockN{Jundong Liu}
  \IEEEauthorblockA{School of EECS \\
    Ohio University\\
    Athens, OH 45701} }


%


\maketitle

\begin{abstract}
  In this paper, we propose a nonlinear distance metric learning
  scheme based on the fusion of component linear metrics. Instead of
  merging displacements at each data point, our model calculates the
  velocities induced by the component transformations, via a geodesic
  interpolation on a Lie transformation group. Such velocities are
  later summed up to produce a global transformation that is
  guaranteed to be diffeomorphic. Consequently, pair-wise distances
  computed this way conform to a smooth and spatially varying metric,
  which can greatly benefit $k$-NN classification. Experiments on
  synthetic and real datasets demonstrate the effectiveness of our
  model.

  \end{abstract}


%
\IEEEpeerreviewmaketitle

\section{Introduction}


In mathematics, a metric is a function that defines distances among
data samples. It plays a crucial role in many machine learning and
data mining algorithms, e.g. $k$-means clustering and $k$-NN
classifier. The Euclidean distance is the most commonly used metric,
which essentially assign all feature components with equal
weights. Learning a customized metric from the training samples to
grant larger weights to more discriminative features can often
significantly improve the performance of associated classifiers
\cite{bellet2013survey,yang2006distance}.

Based on the form of the learned metric, distance metric learning
(DML) solutions can be categorized into linear and nonlinear groups
\cite{yang2006distance}. Linear models focus on estimating a ``best''
affine transformation to deform the data space, such that the resulted
pair-wise distances would very well agree with the supervisory
information brought by training samples. Many early works have
concentrated on linear methods as they are easy to use, convenient to
optimize and robust to overfitting \cite{bellet2013survey}.
However, when applied to process data with nonlinear structures,
linear models show inherently limited separation capability.


Recent years have seen great efforts to generalize linear DML models
for nonlinear cases. Such extensions have been pushed forward mainly
along two directions: {\it kernelization} and {\it localization}.
{\it Kernelization} \cite{torresani2007kLMNN,kwok2003learning} embeds
the input data into a higher dimensional space, hoping that the
transformed samples would be more separable under the new domain.
While effective in extending some linear DML models, kernelization
solutions are prone to overfitting \cite{bellet2013survey}, and their
utilization is inherently limited by the sizes of the kernel matrices
\cite{he2013kernel}.

{\it Localization} focuses on forming a global distance function
through the integration of multiple local metrics estimated within
local neighborhoods or class memberships.
{\it How, where} and {\it when} the local metrics are computed and
blended are the major issues \cite{ramanan2011} of this
procedure. Straightforward piecewise linearization has been
utilized on several global linear DML solutions, including NCA
\cite{NCA} and LMNN \cite{Weinberger06distance}, to develop their
nonlinear versions (msNCA \cite{msNCA} and mm-LMNN
\cite{weinberger2009distance}). The nonlinear metrics in GLML
\cite{noh2010generative} and PLML \cite{wang2012parametric} are
globally learned, on top of basis neighborhood metrics obtained
through the minimization of the nearest neighbor (NN) classification
errors.
To alleviate overfitting and impose general regularity across the
learned metrics, metric/matrix regularization has been employed in
several solutions \cite{msNCA,noh2010generative,wang2012parametric}.

With the exception of PLML, most localization solutions combine
component metrics in a rather primitive manner --- usually the
distance between a pair of data samples is computed as the geodesic
going through multiple classes/neighborhoods, with no smoothing
control around the neighborhood boundaries. Theoretic discussions are
generally lacking as to how the sharp changes across the boundaries
would affect the overall classification.

Solutions that directly integrate nonlinear space transformations with
classifiers have also been proposed. Shi {\it et al.} combines
Thin-plate spline (TPS) with $k$-NN and SVM \cite{shi2015nonlinear,
  chen2015nonlinear, shi2017nonlinear} and Zhang {\it et al.}
\cite{Zhang_ICMLA2016, Zhang_MICCAI2017} utilize coherent point
drifting (CPD) models with Laplacian SVM for semi-supervised
learning. Significant improvements over the corresponding linear
models have been reported.

Integrating DML with deep learning models, especially convolutional
neural networks (CNNs), has been attempted in several latest studies
\cite{hu2014discriminative, hu2015deep,
  sun2014deep,yi2014deep,hoffer2015deep, han2015matchnet}.  In
\cite{sun2014deep,yi2014deep,hoffer2015deep, han2015matchnet}, CNNs
are utilized to replace the ``hand-craft'' feature engineering step,
and different architectures have been explored to fully optimize the
overall deep metric learning networks.  Although many of the deep DML
models produce state-of-the-art results, they require a large amount
of training data, as a prerequisite, to perform effectively.

In this paper, we propose a novel piecewise linearization for local
metrics fusion. Unlike most linearization DML methods, 
our model carries out metric merging based on the velocities of
individual transformations at each data point. Such velocities are
generated through a geodesic interpolation, on a Lie group, between
the identity transformation and the target transformation. The
resulted overall nonlinear motion
is guaranteed to be a diffeomorphism (a bijective map that is
invertible and differentiable). We term our solution {\it nonlinear
  metric learning through Geodesic Polylinear Interpolation} (ML-GPI)
model.


The remaining of the paper is organized as follows. Section 2 presents
some preliminaries for metric learning, followed by the description of
LMNN and its nonlinear extension mm-LMNN. Section 3 introduces our
ML-GPI solution, as well as its advantages over the existing
models. Experimental results are presented in section 4, and section
5 concludes this paper. 



\section{Preliminaries}

A metric space is a set $\mathcal{X}$ that has a notion of distance
between every pair of points.  The goal of metric learning is to learn
a ``better'' metric with the aid of the training samples
$\mathbf{x_1}, \mathbf{x_2}, ..., \mathbf{x_n} \in \mathcal{X}$.  Let
the metric to be sought denoted by $D_\mathbf{A}$, controlled by
certain parameter vector $\mathbf{A}$. For the Mahalanobis metric,
$\mathbf{A}$ is a positive semi-definite (PSD) matrix
$\mathbf{M} \in \mathbb{R}^{d \times d}$, where $d$ is the number of
features.  If a metric keeps the same-class pairs close while pushing
those in different classes far away, it would likely be a good
approximation to the underlying semantic metric.

\subsection{Global linear model: LMNN}
LMNN is one of the most widely-used Mahalanobis metric learning
algorithms, and has been the subject of many extensions
\cite{torresani2007kLMNN,park2011efficiently,nguyen2008metric,der2012latent,weinberger2009distance}. Our
ML-GPI model is inspired and implemented based on mm-LMNN, one of the
nonlinear extensions of LMNN.
Therefore, we briefly review these two models here.

Unlike many other global metric learning methods, LMNN defines the
constraints in a local neighborhood, where the ``pull force'' within
the class-equivalent data and the ``push force'' for the
class-nonequivalent data (the "imposters") are optimized to lead a
balanced trade-off. Let $\mathcal{P}$ be the set of same-class pairs,
and $\mathcal{N}$ be that of different-class pairs. Formally, the
constraints used in LMNN are defined in the following way:
\begin{itemize}
\item class-equivalent constraint in a neighborhood:\\
  \text{$\mathcal{P}_{nn} = \{({\mathbf{x}}_i,{\mathbf{x}}_j)|
    ({\mathbf{x}}_i, {\mathbf{x}}_j) \in \mathcal{P};
    {\mathbf{x}}_j \text{ and } {\mathbf{x}}_i
    \text{ are neighbors} \}$};
\item class-nonequivalent constraint in a neighborhood:\\
   \text{$\mathcal{N}_{nn} = \{({\mathbf{x}}_i,{\mathbf{x}}_j)|
    ({\mathbf{x}}_i, {\mathbf{x}}_j) \in \mathcal{N};
    {\mathbf{x}}_j \text{ and } {\mathbf{x}}_i
    \text{ are neighbors} \}$};
\item Relative triplets in a neighborhood:\\
  \text{$\mathcal{T}_{nn} = \{({\mathbf{x}}_i,{\mathbf{x}}_j,
    {\mathbf{x}}_k)| ({\mathbf{x}}_i, {\mathbf{x}}_j) \in
    \mathcal{P}_{nn}; ({\mathbf{x}}_i, {\mathbf{x}}_k)
    \in \mathcal{N}_{nn} \}$}.
\end{itemize}

Then, the Mahalanobis metric is learned through the following convex
objective:
\begin{equation}
\label{eqn:LMNN}
\small
 \begin{aligned}
   & \underset{{\mathbf{M}}}{\text{min}}
   & & J({\mathbf{M}}) =  \displaystyle\sum_{({\mathbf{x}}_i, {\mathbf{x}}_j) \in \mathcal{P}_{nn}} D_{\mathbf{M}}^{2}({\mathbf{x}}_i, {\mathbf{x}}_j) + \mu \displaystyle\sum_{i,j,k} \xi_{ijk}  \\
   & \text{s.t.}
   &  & {\mathbf{M}} \succeq 0,\\
   & & & \xi_{ijk} \geq 0, \\
   & & & D_{\mathbf{M}}^{2}({\mathbf{x}}_i, {\mathbf{x}}_k) -
   D_{\mathbf{M}}^{2}({\mathbf{x}}_i, {\mathbf{x}}_j) \geq 1 -
   \xi_{ijk},\quad \forall ({\mathbf{x}}_i, {\mathbf{x}}_j,
   {\mathbf{x}}_k) \in \mathcal{T}_{nn} \nonumber
 \end{aligned}
\end{equation}

where $\mu \in [0,1]$ controls the ``pull/push'' trade-off. A tailored
numerical solver based on gradient descent and book-keeping strategy
is utilized, enabling LMNN to perform efficiently in practice.

\subsection{Nonlinear extension through piecewise linearization}

To achieve nonlinear extension of linear metric learning models,
piecewise linearization is a simple yet popular solution, which is
commonly chosen by per-class methods
\cite{weinberger2009distance,msNCA,wang2012parametric,ramanan2011}. Taking
mm-LMNN and PLML as examples, the idea of piecewise linearization is
rather simple: in order to learn separate Mahalanobis metrics in
different parts of the data space, the original training data are
firstly partitioned into $c$ disjoint clusters based on either spatial
$k$-means or class labels. Different metrics
${\mathbf{M}}_{b_1}, {\mathbf{M}}_{b_2}, \dots, {\mathbf{M}}_{b_c}$ at
the cluster centers
${\mathbf{U}}_1, {\mathbf{U}}_2, \dots, {\mathbf{U}}_c$ are learned
simultaneously.

\begin{figure}
\begin{center}
\includegraphics[width=2.5in]{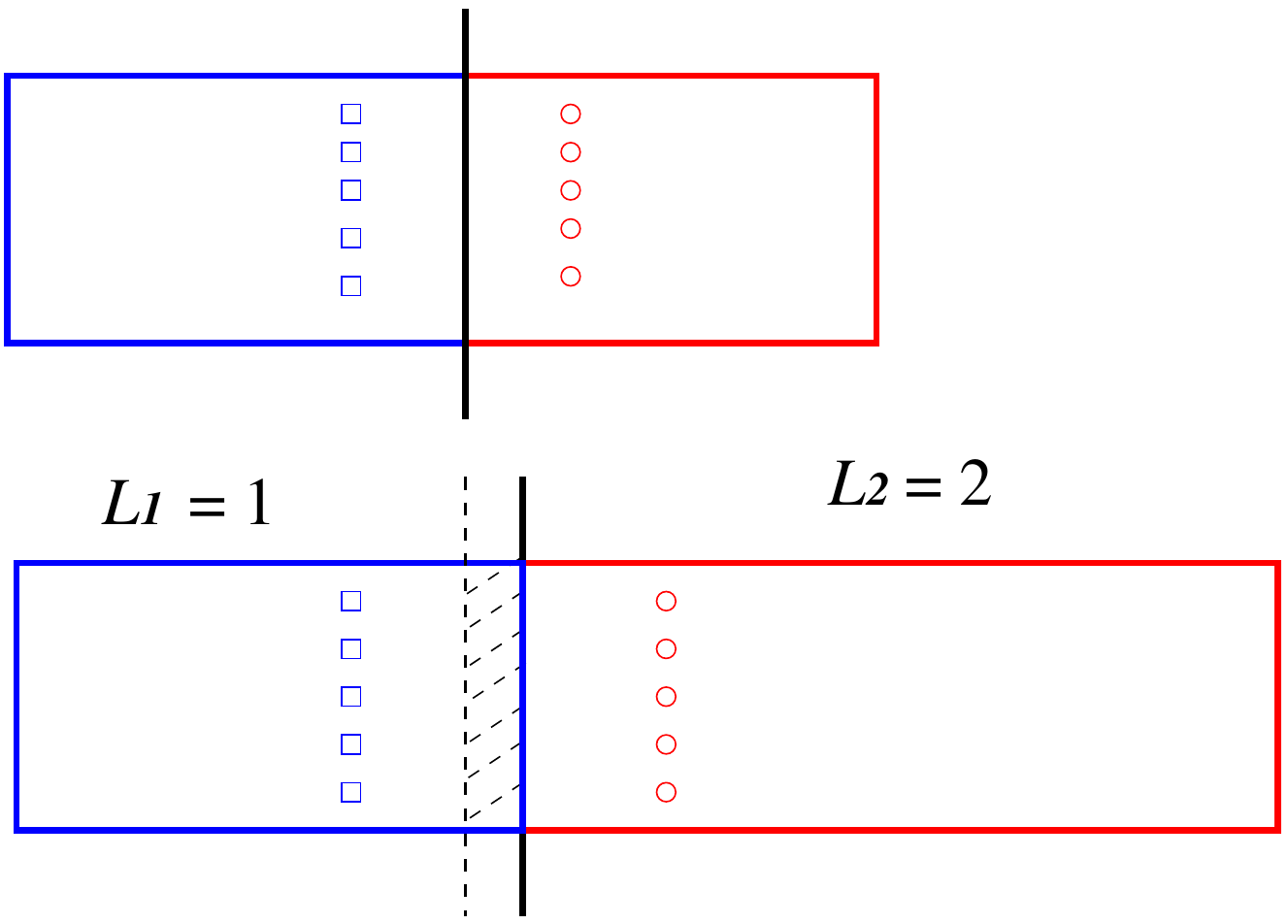}
\end{center}
\caption{mm-LMNN: artificial shift of decision boundary caused by
  unbalanced component metrics.}
 \label{fig:shift}
\end{figure}

In mm-LMNN, data points at the testing stage are mapped with different
component metrics before the $k$-NN classification decision is
made. While straightforward and generally effective, this scheme has
the tendency to artificially displace the decision boundaries,
especially when the component metrics are with rather different
scales. 
This effect is illustrated in Fig.~\ref{fig:shift}. The first row
shows two classes of points, red and blue, as well as the associated
$k$-NN decision boundary (figuratively speaking, not the exact
boundary), which is roughly the middle line separating two boundary
group points. When the data space is transformed with unbalanced
component transformations on the two halves, e.g., ${\mathbf{L}}_1$ is the
identity transformation while ${\mathbf{L}}_2$ doubles the horizontal dimension
of the red side, the samples around the original decision boundary
(shadow area) will be classified into the blue class, as their
distances to the red circles have been doubled. This means the new
decision boundary (the new middle line) is artificially shifted due to
the unbalanced metrics. Test samples, if falling into the shadow area,
will be misclassified.

In PLML, at each instance ${\mathbf{x}}_i$, the local metric ${\mathbf{M}}_i$ is
parameterized through weighted linear combination
\cite{wang2012parametric}:
\begin{equation}
  {\mathbf{M}}_i = \sum_{b_k} W_{ib_k} {\mathbf{M}}_{b_k}, \ \ \ W_{ib_k} \geq 0, \ \ \  \sum_{b_k} W_{ib_k} = 1
\label{wmetric}
 \end{equation}

\begin{figure}
\begin{center}
\includegraphics[width=1.6in]{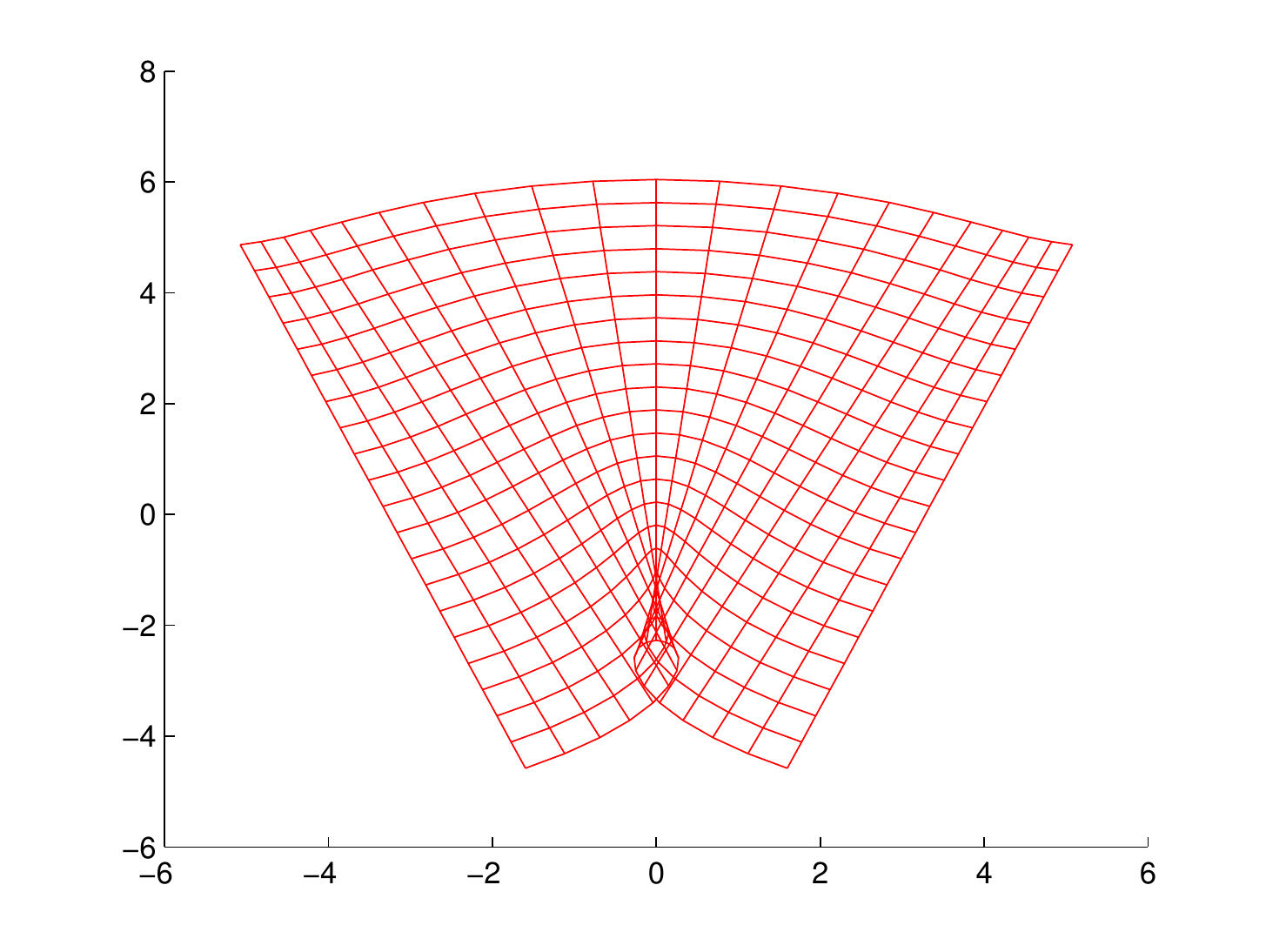}
\includegraphics[width=1.6in]{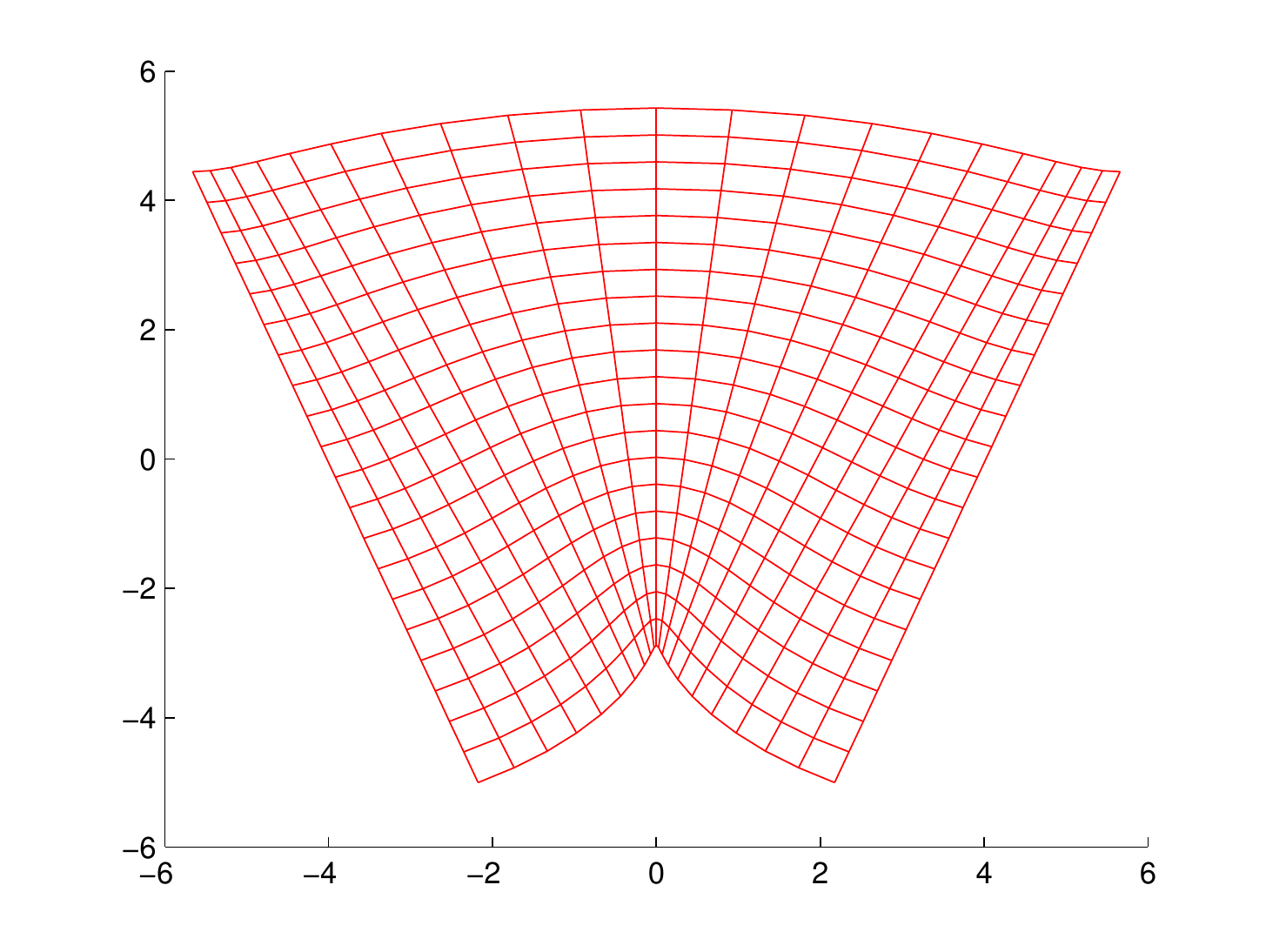}
\end{center}
\caption{Combination through displacements (left) vs. combination
  through velocities (right).}
\label{fig:combination}
\end{figure}

 $W_{ib_k}$ is the weight of the cluster metric ${\mathbf{M}}_{b_k}$ for the
 instance ${\mathbf{x}}_i$. Using Eqn.(\ref{wmetric}), the squared distance of
 ${\mathbf{x}}_i$ to ${\mathbf{x}}_j$ is:
\begin{equation}
 d^2 ({\mathbf{x}}_i, {\mathbf{x}}_j) = \sum_{b_k} W_{ib_k}  d^2_{{\mathbf{M}}_{b_k}} ({\mathbf{x}}_i, {\mathbf{x}}_j)
\label{wdistance}
 \end{equation}

 where $d^2_{{\mathbf{M}}_{b_k}} ({\mathbf{x}}_i, {\mathbf{x}}_j)$ is the squared Mahalanobis distance
 between ${\mathbf{x}}_i$ and ${\mathbf{x}}_j$ under the cluster metric ${{\mathbf{M}}_{b_k}}$. In
 other words, pair-wise distances, as well as the combination of
 component metrics, is conducted through weighted averaging of the
 associated {\it displacements}. In principle, this simple strategy
 could be applied to any Mahalanobis metric learning algorithms,
 extending a global solution to solve nonlinear cases. However, this
 approach has an inherent drawback: although all the component metrics
 are invertible, the resulting global metric is not necessarily
 invertible in general.
 Fig.~\ref{fig:combination} (left) shows the direct fusion of two
 estimated linear metrics, where a folding around the boundary area
 occurs. Foldings in space indicate that two different data points
 could be mapped to a same point after transformation, which would
 generate inconsistent classification decisions.

\section{ML-GPI: nonlinear metric learning through geodesic polylinear
  interpolation}

As we explain in the previous section, to merge local metrics with
different scales, the decision boundaries between classes could be
artificially displaced, leading to erratic classification results.  A
smooth transition could provide a remedy. PLML model takes the
smoothness issue into consideration, and imposes a manifold
regularization step to ensure local metrics to vary smoothly across
the entire data domain. However, the distance between each data pair
is obtained through weighted summations over the individual distances
for involved bases. While the weights are smooth, such distance
summation does not ensure diffeomorphic and smoothness in the overall
distance field.


Our solution for smooth transitions is based on a Lie group geodesic
interpolation approach that has been utilized in motion interpolation
\cite{alexa2002linear,rossignac2011steady} and image registration
\cite{arsigny2005polyrigid,arsigny2009fast} research. With the
component linear metrics, we rely on the averaging of infinitesimal
displacements, or velocities, to generate the combined
transformation. Each transformation ${\mathbf{L}}_k$ is modeled as an
instance in a high dimensional Lie group, and motion velocity induced
by ${\mathbf{L}}_k$ can be calculated through a constant speed
interpolation from the identity transformation ${\mathbf{I}}$ to
${\mathbf{L}}_k$. With the weighted velocity at each data point, the
global metric is subsequently obtained by integrating an Ordinary
Differential Equation (ODE). The result is guaranteed for
invertibility and smoothness \cite{arsigny2009fast}.
Fig.~\ref{fig:combination} (right) shows the fusion of the same linear
metrics (mentioned in last section), through velocity approach.  The
resulted transformation does not fold and remains invertible, which is
in a great contrast with the fusion through displacements, as in
Fig.~\ref{fig:combination} (left).

The overall procedure of our ML-GPI can be decomposed into four steps:

\begin{itemize}
\item Step 1: Derive velocity vectors for individual linear
  metrics. Let $q$ be the total number of component metrics. For
  component metrics
  ${\mathbf{M}}_k = {\mathbf{L}}_k^T{\mathbf{L}}_k (k = 1, 2, ...q)$
  at a data point ${\mathbf{x}}$, a family of velocity vector fields
  ${\mathbf{V}}_k ({\mathbf{x}}, t)$ parameterized by a time parameter
  $t$ ($ 0 \leq t \leq 1$), need to be derived.
  ${\mathbf{V}}_k ({\mathbf{x}}, t)$ represents the velocity incurred
  by ${\mathbf{L}}_k$, therefore it should satisfy a consistency
  property: when integrated from time $0$ and $1$, the accumulated
  transformation should start at the identity transformation
  ${\mathbf{I}}$ and end at ${\mathbf{L}}_k$.
 
\item Step 2: Fuse the velocity vectors in step 1 according to a
  weighting function
  $w_i(x)$. 
  \begin{equation} {\mathbf{x}}' = \mathbf{V}({\mathbf{x}}, t) =
    \sum_k w_k({\mathbf{x}}) \mathbf{V}_k({\mathbf{x}}, t)
\label{wspeed}
\end{equation}

\item Step 3: Integration along the velocity ODE. Under
  this velocity framework, the motion destination of each point
  {${\mathbf{x}}_{i_0}$} through the combined global
  transformation ${\mathbf{L}}$ is obtained via the integration of
  Eqn.~(\ref{wspeed}) between $t=0$ and $t=1$, with the initial
  condition ${\mathbf{x}}_i(0) = {\mathbf{x}}_{i_0}$.

\item Step 4: The distance of two data point ${\mathbf{x}}_i$ to
  ${\mathbf{x}}_j$ through the global transformation ${\mathbf{L}}$
  will be the Euclidean distance between their respective
  destinations.

\end{itemize} 

\begin{figure}
\begin{center}
  \includegraphics[width=2.8in]{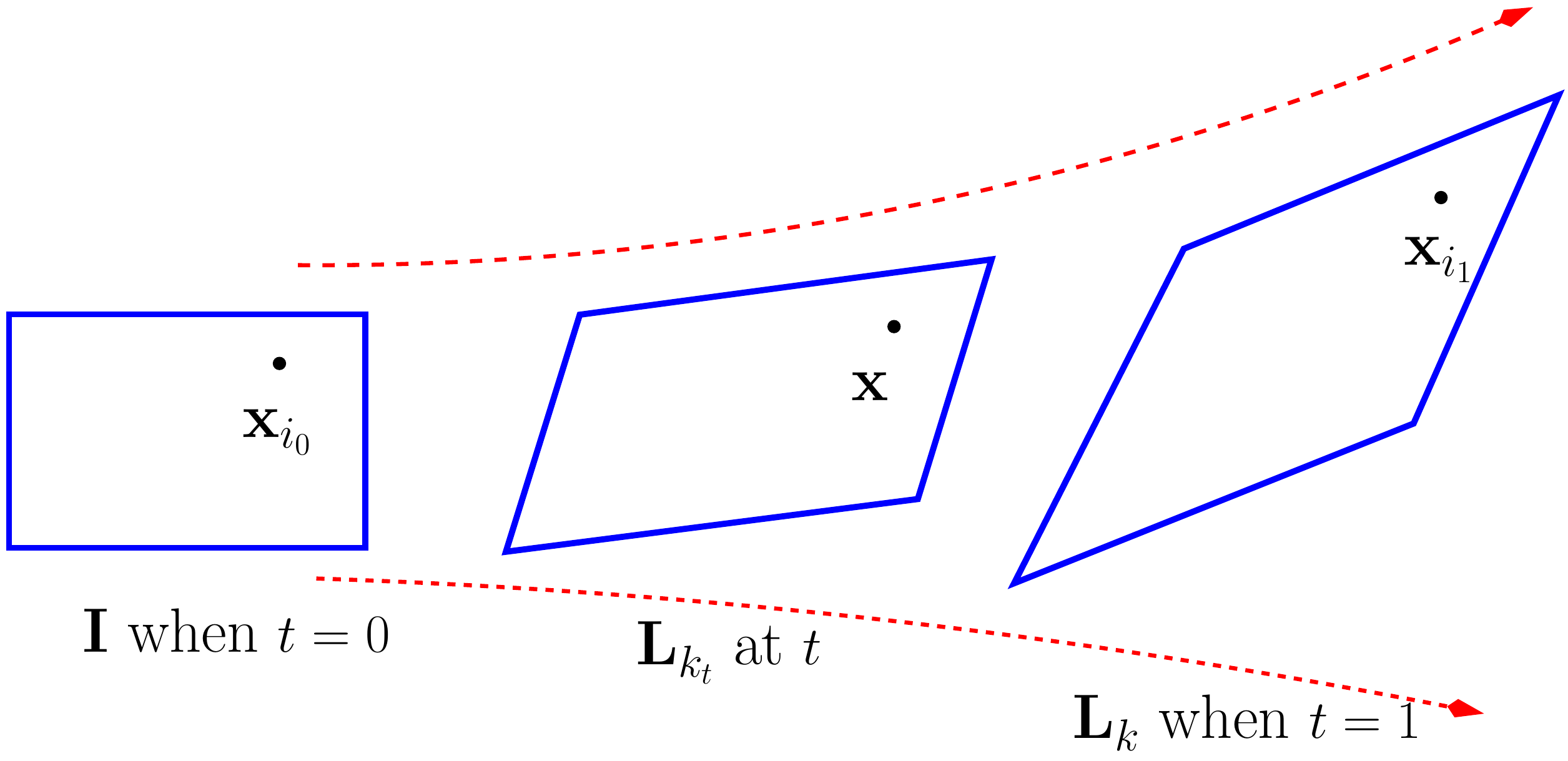}
\end{center}
\caption{An illustration of transformation interpolation for
  point-wise velocities. Linear transformation ${\mathbf{L}}_k$ is the
  destination when $t=1$. We seek to estimate a sequence of
  intermediate transformations ${\mathbf{L}}_{k_t}$ at $t \in (0, 1)$ as well as
  the motion velocity $V({\mathbf{x}}, t)$ at location ${\mathbf{x}}$. }
\label{fig:Lie_interpolation}
\end{figure}

\subsection{Velocity vectors and weighting function}


The mapping of points ${\mathbf{x}}_{i_0}$ to
${\mathbf{x}}_{i_1} = {\mathbf{L}}_k \cdot {\mathbf{x}}_{i_0}$ through
a linear transformation ${\mathbf{L}}_k$ can be interpreted in
infinitely many different ways. One “reasonable” path is to regard the
transformations as instances in a Lie transformation group, and carry
out the morphing along the geodesic from the identity matrix
${\mathbf{I}}$ (starting transformation, at $t = 0$) to
${\mathbf{L}}_k$ (destination transformation at $t=1$). Within this
morphing procedure, each point ${\mathbf{x}}$ is moving with certain a
velocity $\mathbf{V}({\mathbf{x}}, t)$.
An illustration of this interpolation scheme is given in Fig.~\ref{fig:Lie_interpolation}. 

As many metric learning solutions, including LMNN and mm-LMNN, output
real invertible matrices, it is justified to only consider component
transformations ${\mathbf{L}}_k$ in the general linear group
{$\mathbf{GL}$}({$n$, {$\mathbf R$}). As a manifold,
  {$\mathbf{GL}$}({$n$, {$\mathbf R$}) is not connected but rather has
    two connected components: the matrices with positive determinant
    ({$\mathbf{GL}$}$_{+}$({$n$, {$\mathbf R$})) and the ones with
      negative determinant ({$\mathbf{GL}$}$_{-}$({$n$, { R})). Since
        the identity matrix ${\mathbf{I}}$ is in
        {$\mathbf{GL}$}$_{+}$, we specify the transformation group to
        be {$\mathbf{GL}$}$_{+}$({$n$, {$\mathbf R$}) and only
          consider the matrices with positive determinants.

          For a Lie group $G$, consider two element matrices
          $a, b \in G$. We desire to find an interpolation between the
          two elements, according to a time parameter $t \in [0,
          1]$. Define a function that will perform the interpolation:
\begin{eqnarray}
  f: G \times G \times R & \rightarrow & G  \nonumber \\
  f(a, b, 0) & = & a \nonumber \\
  f(a, b, 1) & = & b \nonumber
\end{eqnarray}
The function can be obtained by transforming the interpolation
operation into the tangent space at the identity $T_eG$, performing
a linear combination there, and then transforming the resulting
tangent vector back onto the manifold. First consider the group
element that takes $a$ to $b$:
\begin{eqnarray}
 d & \equiv & b \cdot a^{-1}  \nonumber \\
 d \cdot a & = & b \nonumber
\end{eqnarray}

Now compute the corresponding Lie algebra vector and assume the motion
is with a constant speed on the tangent space $T_eG$:
\begin{eqnarray}
 d(t) & = & t \cdot {{\mathbf{log}}}(d)
\end{eqnarray}

Then transform back into the manifold using the exponential map,
yielding an intermediate transformation at time $t$:
\begin{eqnarray}
 d_t & = & {\mathbf{exp}}(d(t))
\end{eqnarray}

Combining these three steps leads to a solution for $f$:
\begin{eqnarray}
  f(a, b, t) & = & d_t \cdot a   \nonumber \\
             & = & {\mathbf{exp}}(t \cdot {\mathbf{log}}(b \cdot a^{-1})) \cdot a \nonumber
\end{eqnarray} 

Note that the intermediate transformation is always on the manifold,
due to the operation of the exponential map.

\begin{figure}
\begin{center}
\includegraphics[width=2.1in]{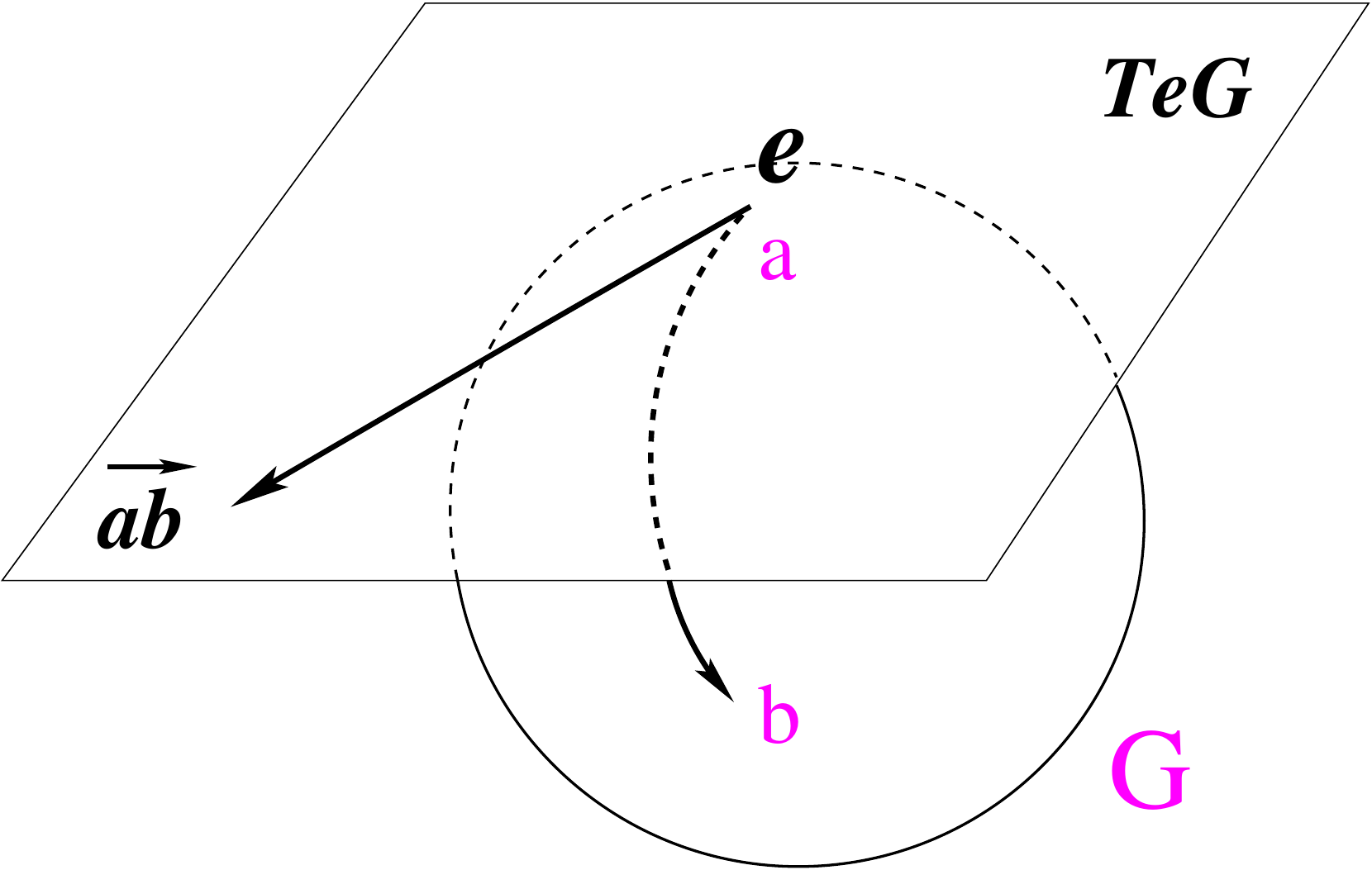}
\end{center}
\caption{Illustration of the exponential and log maps for the
  transformation Lie group. }
\label{fig:expmap}
\end{figure}

Step 1 of our ML-GPI model is a straightforward application of the
above derivations. Our goal is to estimate the velocity
$\mathbf{V}_k({\mathbf{x}}, t)$ incurred by transform ${\mathbf{L}}_k$
at point ${\mathbf{x}}$ and time $t$. The transformation travels with
constant speed on {$\mathbf{GL}$}$_{+}$({$n$, {$\mathbf R$}) through
  the geodesic from identity matrix ${\mathbf{I}}$ $(t=0)$ to
  ${\mathbf{L}}_k$ $(t=1)$, so $a = {\mathbf{I}}$ and
  $b = {\mathbf{L}}_k$ in the above derivations. For a point
  transformed to ${\mathbf{x}}$ after time $t$ through an intermediate
  transformation ${{\mathbf{L}}}_{k_t}$, let ${\mathbf{x}}_{i_0}$ be
  the original location. Since ${{\mathbf{L}}}_{k_t}$ is a linear
  transformation,
\begin{equation}
  {\mathbf{x}} = {{\mathbf{L}}}_{k_t} \cdot {\mathbf{x}}_{i_0} = f({\mathbf{I}}, {\mathbf{L}}_k, t) \cdot {\mathbf{x}}_{i_0} = {\mathbf{exp}}(t \cdot {\mathbf{log}}({\mathbf{L}}_k)) \cdot {\mathbf{x}}_{i_0}  
\end{equation}
Calculate the derivative of ${\mathbf{x}}$ w.r.t. to $t$, we obtain
the velocity:
\begin{equation} {\mathbf{V}}_k({\mathbf{x}}, t) =
  {\mathbf{log}}({\mathbf{L}}_k) \cdot {\mathbf{exp}}(t \cdot
  {\mathbf{log}}({\mathbf{L}}_k)) \cdot {\mathbf{x}}_{i_0} \ \ \ \ \ t
  \in [0, 1]
\end{equation}


{\bf Maintain matrices in {$\mathbf{GL}$}$_{+}$} A matrix with
negative determinant would flip the data space. Most global linear
metric learning algorithms allow the estimated linear transformation
to be in {$\mathbf{GL}$}$_{-}$({$n$, {$\mathbf R$}), as a flipping
  does not affect the classification that follows. For piecewise
  linear models, however, flippings could impose a serious problem. When
  neighboring metrics fall in {$\mathbf{GL}$}$_{+}$({$n$,
    {$\mathbf R$}) and {$\mathbf{GL}$}$_{-}$({$n$, {$\mathbf R$})
      respectively, which are disconnected subgroups of
      {$\mathbf{GL}$}({$n$, {$\mathbf R$}), merging or averaging such
        matrices would lead to disastrous results. In our ML-GPI
        model, we specify all the component metrics ${\mathbf{L}}_k$
        to be in {$\mathbf{GL}$}$_{+}$({$n$, {$\mathbf R$}).
          To this end, we modified mm-LMNN and adopted a procedure
          similar to the projected gradient approach utilized in
          \cite{xing2003distance}. At each iteration of mm-LMNN, we
          check the determinant of the estimated transformations
          ${\mathbf{L}}_k$. If any of them falls in
          {$\mathbf{GL}$}$_{-}$({$n$, {$\mathbf R$}), we project it
            back to {$\mathbf{GL}$}$_{+}$({$n$, {$\mathbf R$}) by
              changing the sign of one of the Jordan eigenvalues.

{\bf Weight functions} With the velocities estimated from individual
metrics, combination can be conducted through certain weighting
function. Weight functions model influence in space of each component
metric, and partly control the sharpness of transitions among the
fused linear transformations.  A desired weight function should ensure
a smooth transition across class/region boundaries. In this work, we
utilize radial distance functions for such control, where the
influence of each transformation is gradually reduced as the distance
away from the class center grows. Let $\vec{c_i}$ be the computed
class center (group mean in spatial coordinates). The weight functions
we choose in all the experiments of this paper take the form of
$w_i({\mathbf{x}})=1/(1+(||{\mathbf{x}}-{\mathbf{c}}_i||)/\sigma)^2))$.
$\sigma$ serves as an attenuation constant that controls the rate of
influence reduction, and it can be roughly set to the radius of the
data points in the same class. $w_i$ is then normalized as
$w_i=w_i/(\sum
w_i)$. 

Fig.~\ref{fig:velocities} shows an example of metric fusion through
our ML-GPI model. The first row are two component linear
transformations: rotations of opposite angles of magnitude $0.63$
radians around the centers of the respective regions.
Fig.~\ref{fig:velocities}.c shows the combined point-wise velocities
estimated in Step 1, and the overall transformation computed through
ML-GPI is shown in Fig.~\ref{fig:velocities}.d. As evident, the
combined transformation field is smooth and invertible.

\begin{figure}[t]
\centering
\begin{tabular}{cc}
{\includegraphics[width=1.6in]{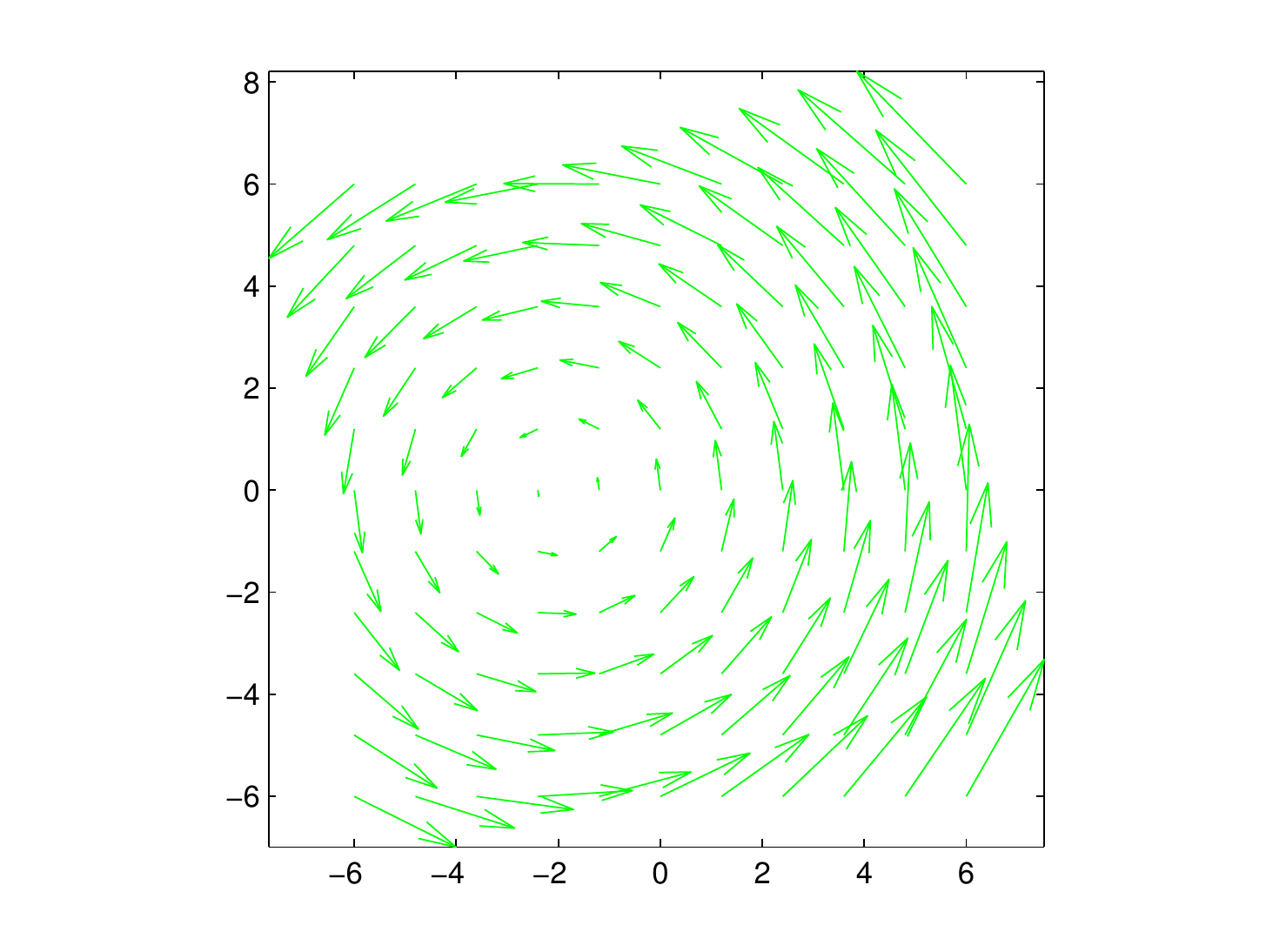}} & {\includegraphics[width=1.6in]{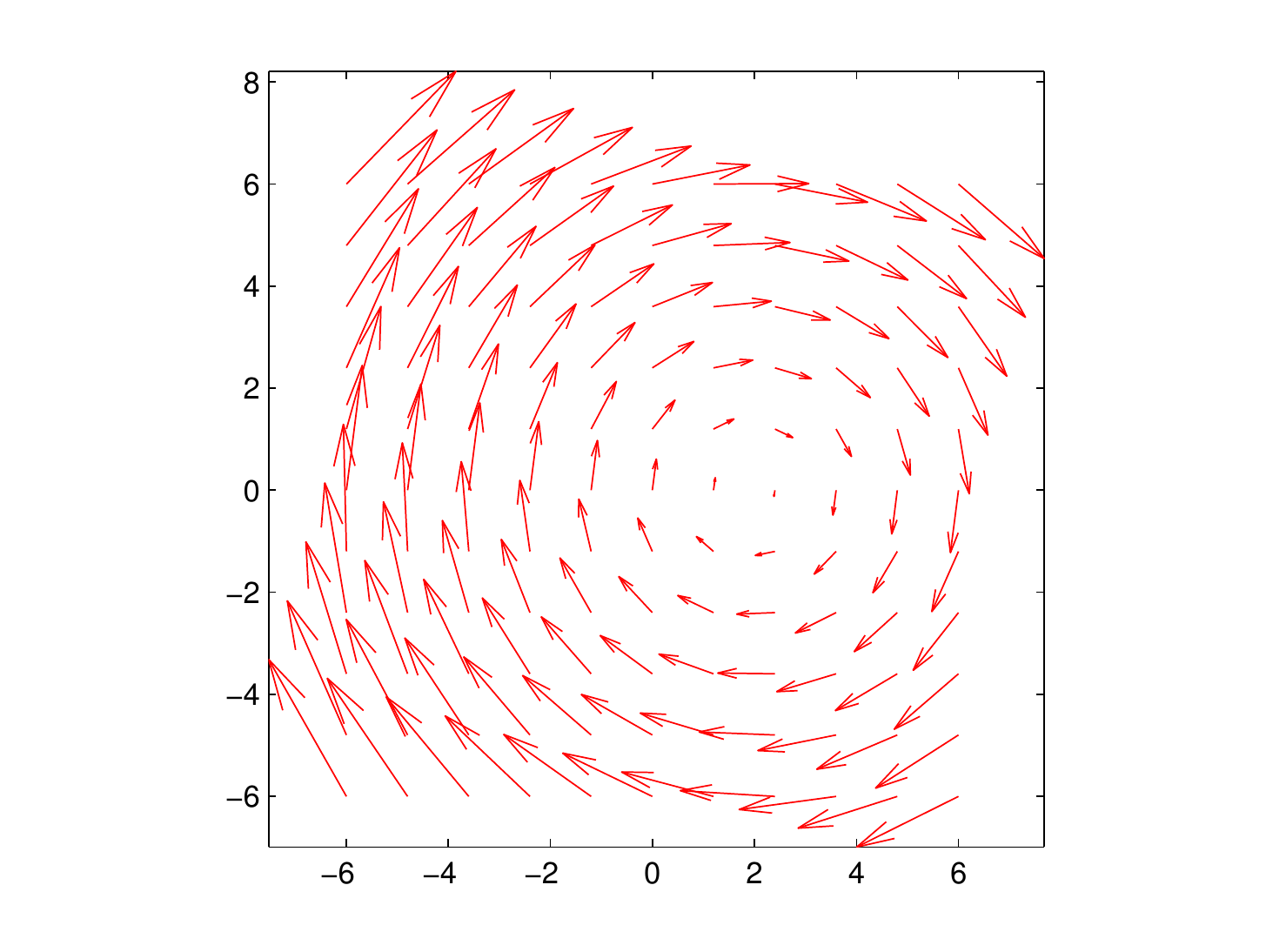}} \\
(a) Linear metric 1 & (b) Linear metric 2  \\
{\includegraphics[width=1.6in]{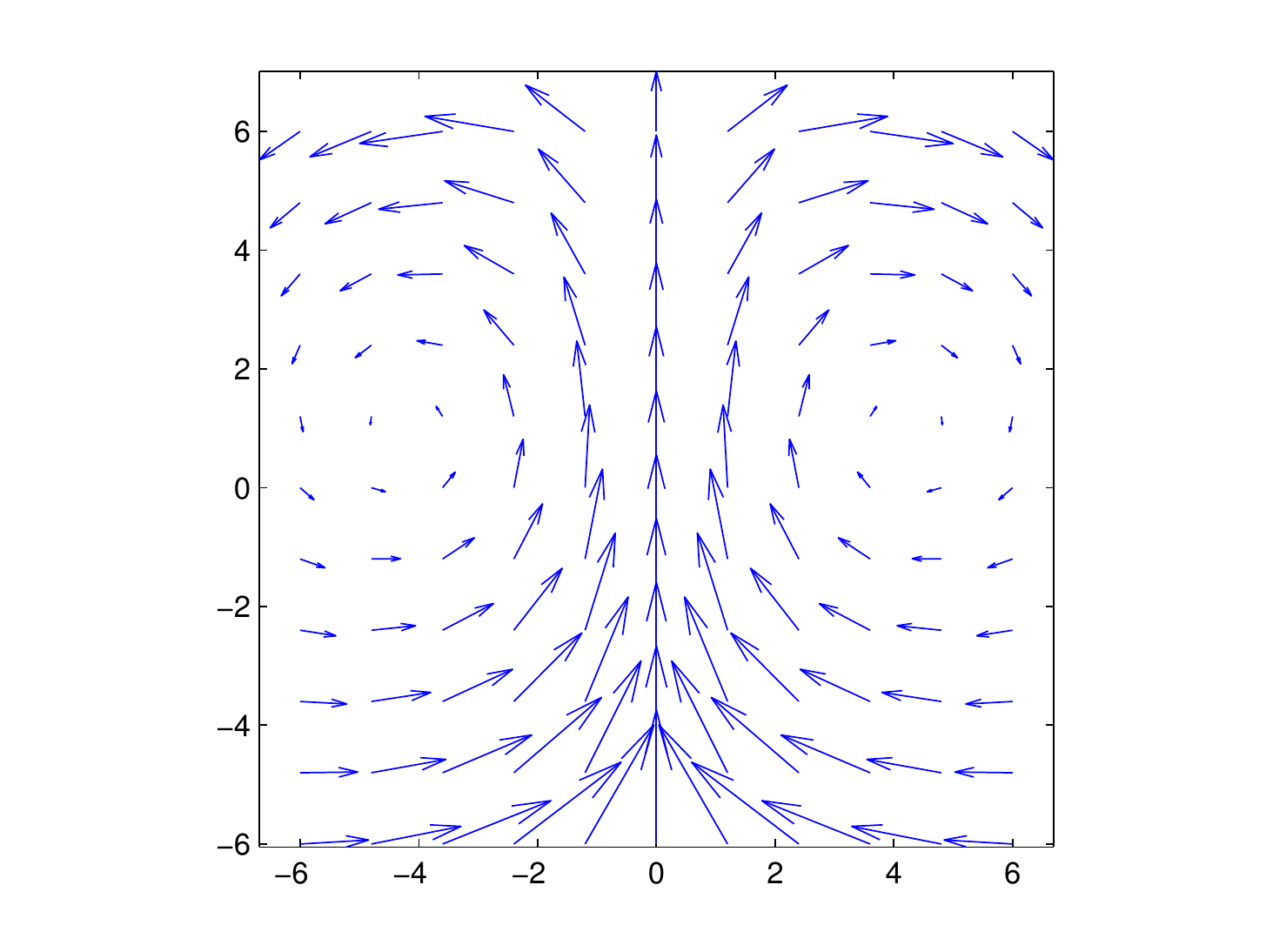}} &  {\includegraphics[width=1.6in]{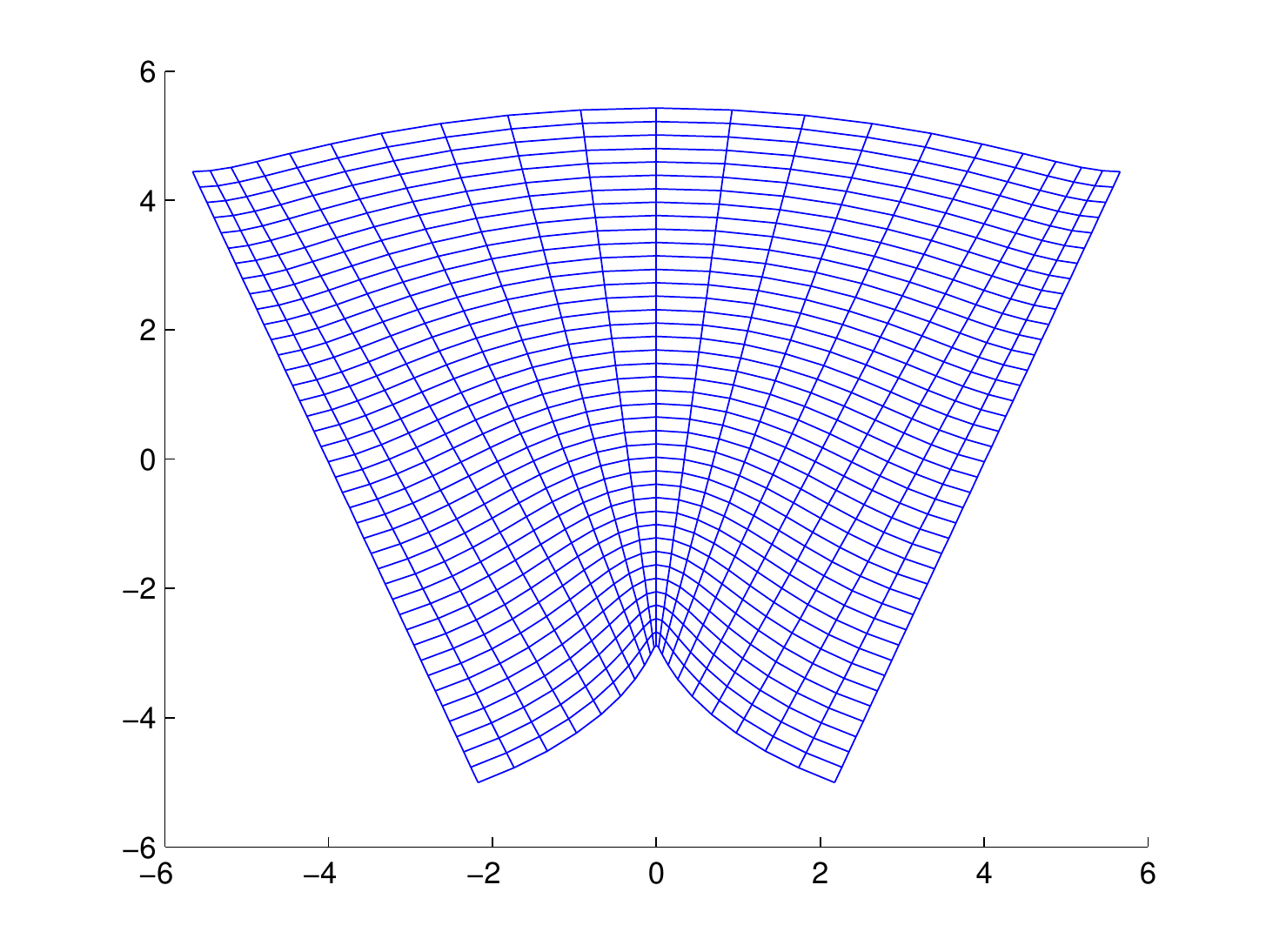}} \\
(c) Combined velocities & (d) Overall transformation  \\
\end{tabular}
\caption{Velocities generated from linear transformations, and fusion
  of the velocities.}
\label{fig:velocities}
\end{figure}



\section{Experiments and Results}
In this section, we present evaluation and comparison results of
applying our proposed ML-GPI nonlinear DML methods on both synthetic
and real datasets.

\subsection{Synthetic data: effects on decision boundaries}


As mentioned before, piecewise linearization strategies commonly
result in boundary shifts in classification due to the lack of
smooth transitions across class boundaries. Our model, ML-GPI,
ensuring a diffeomorphic transformation, can avoid this problem.

To verify this claim, an experiment with synthetic data is designed as
shown in Fig~\ref{fig:decision-polyliner}.  Two classes of samples are
distributed in the original space with a ``stripe'' type.  Compared to
``stripes'' in Class 2, the distribution density in Class 1 is lower,
which might decrease the ``pulling force'' within Class 1.  This
imbalance may result in boundary shifts between classes.

Fig.\ref{fig:2class-motion} shows the two component transformations
generated in ML-GPI and how they are fused. The two arrows indicate
the general directions of the forces generated by the two affine
transformations. With the weighted summation of the instantaneous
velocities, the deformation across the entire data domain is smooth,
which ensure the class boundary to be updated with spatial
consistency.

The resulted separating lines from mm-LMNN and our ML-GPI model are
shown in Fig.~\ref{fig:decision-polyliner}. It is clear that the
$k$-NN boundary determined by ML-GPI is closer to the ideal one, and
has no displacement (erosion) into Class 2 area as mm-LMNN does, which
also implies that the classification rate by ML-GPI will be higher
than mm-LMNN.

\begin{figure}
\begin{center}
\includegraphics[width=3.0in]{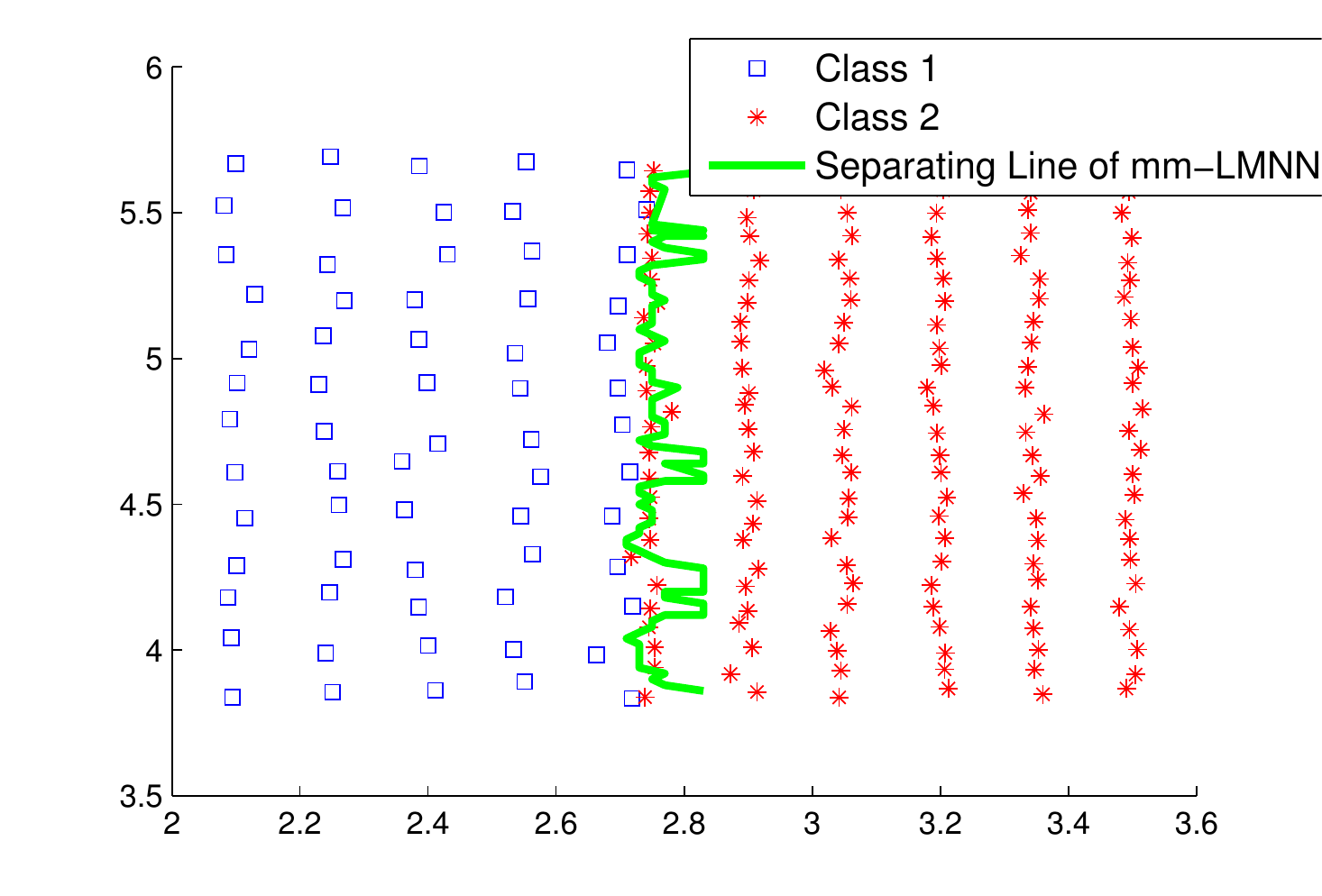}
\includegraphics[width=3.0in]{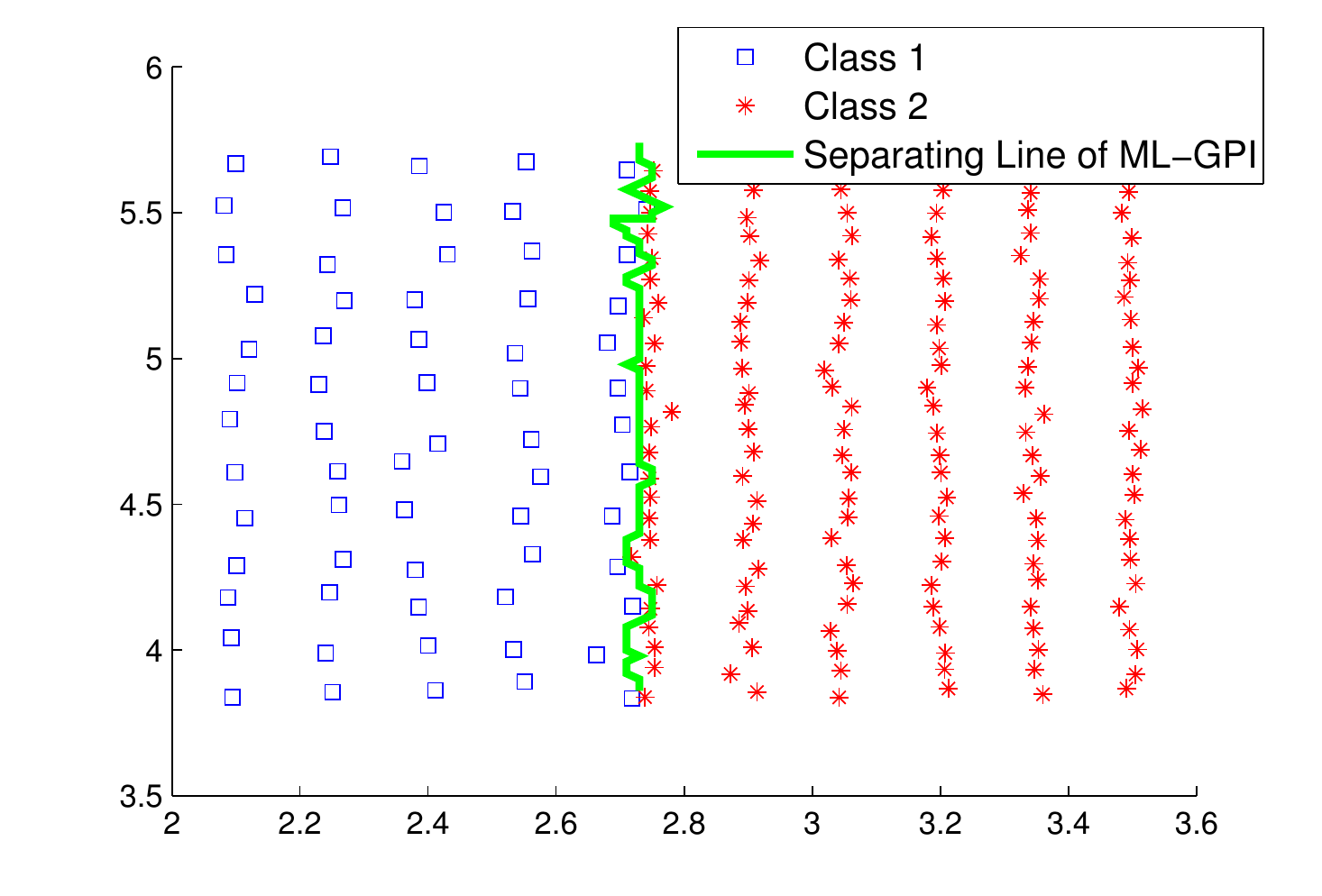}
\end{center}
\caption{The separating lines at the boundary by: mm-LMNN (top) and
  ML-GPI (bottom).}
\label{fig:decision-polyliner}
\end{figure}

\begin{figure}
\begin{center}
\includegraphics[width=3.0in]{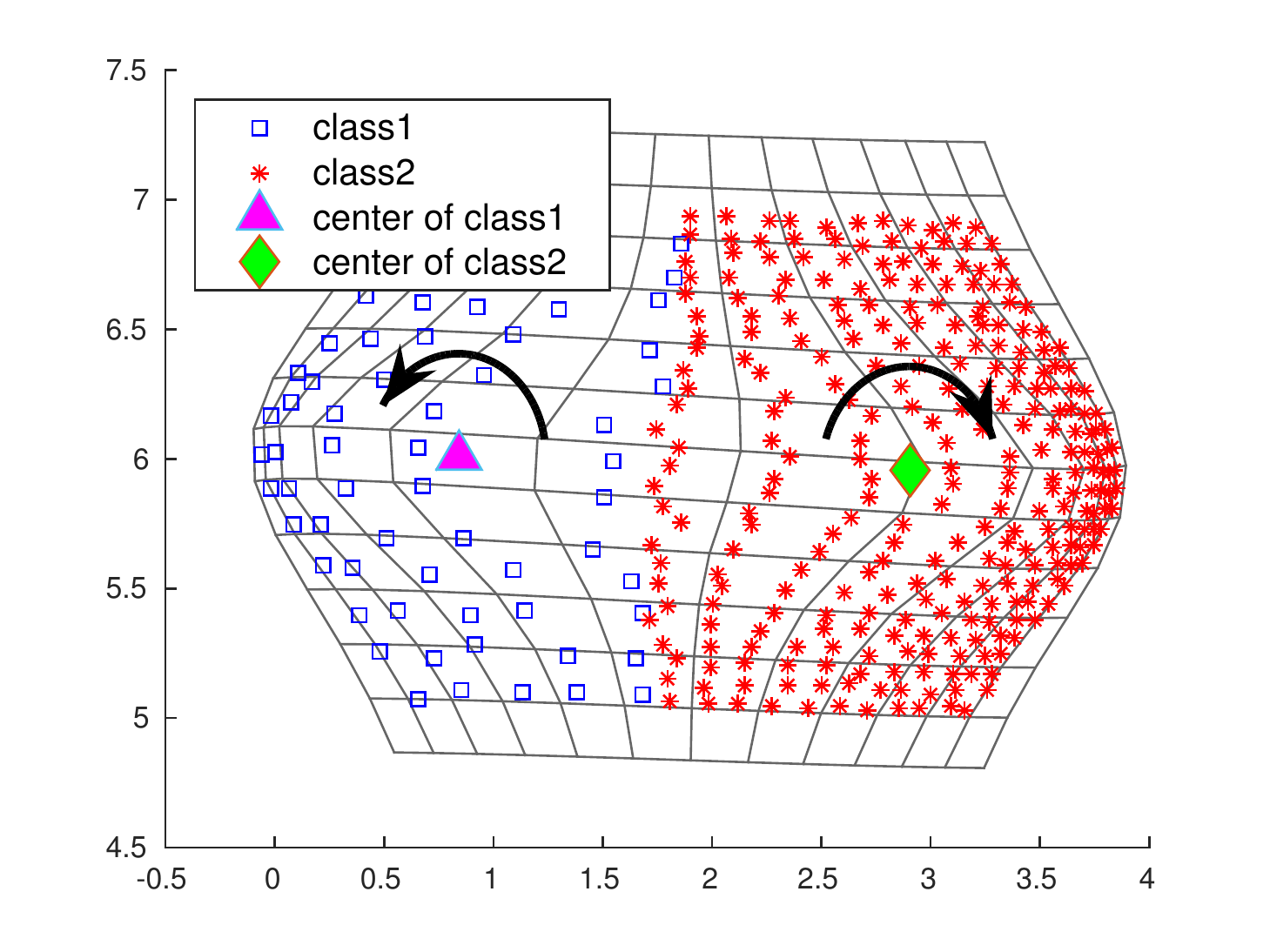}
\end{center}
\caption{Transition of space via ML-GPI}
\label{fig:2class-motion}
\end{figure}

\subsection{Real data: application to Alzheimer's Disease (AD) staging}

Alzheimer's disease (AD) and its early stage, mild cognitive
impairment (MCI), is a serious threat to more than five million
elderly people in US. Identifying intermediate biomarkers of AD is of
great importance for diagnosis and prognosis of the disease. We apply
our proposed ML-GPI model on AD staging problem to demonstrate the
practical usefulness of our algorithm.

321 subjects from the Alzheimer's Disease Neuroimaging Initiative
cohort (56 AD, 104 MCI, and 161 normal controls) were used as the
input data. Sixteen features, including both volume and shape
information for several important subcortical structures are
extracted, and ranked.  To better compare the performance, as well
as identify the effectiveness of the models,
we focus on a much simplified feature set that consists of the first
and second most discriminative features (the volume information of
{\it Hippocampus} and {\it Entorhinal}).

To evaluate our proposed ML-GPI model for the AD staging problem, the
ternary classification experiment (to separate AD/MCI/NC
simultaneously) were conducted, with comparison with mm-LMNN. A
leave-10\%-out 10-fold cross-validation paradigm is adopted for each
model. The best, worst and average classification rates from the 10
validations were computed and included in Table~\ref{T:AD}.  Our
ML-GPI performed much better than mm-LMNN on all three  measurements.
Nevertheless, we want to point out that even though the absolute
values for all of the performance measures obtained by ML-GPI and
mm-LMNN are relatively low (which can be improved later with refined
feature extraction steps, such as more accurate Hippocampus atrophy
estimation), the relative improvements made by ML-GPI over mm-LMNN
could still be an indirect indication on the better performance
achieved by ML-GPI.

\begin{table}[!ht]
\small
\caption{Results on AD
staging problem}
\centering
\begin{tabular}{c|c|ccc|}
  \hline
  \hline
\multicolumn{1}{c|}{}&\multicolumn{1}{c|}{}&
\multicolumn{3}{c}{\textbf{Results}}\\ 
\cline{3-5}
\multicolumn{1}{c|}{\textbf{Classifier}} &\multicolumn{1}{c|}{\textbf{DML Method}}&
\multicolumn{1}{c}{\textbf{Mean}} &
\multicolumn{1}{c}{\textbf{Max}} &
\multicolumn{1}{c}{\textbf{Min}} \\ 
\hline
  \multicolumn{1}{c|}{} 
  & \multicolumn{1}{c|}{\text{mm-LMNN}} & 0.3738$\pm$0.0766 & 0.5152 & \multicolumn{1}{c}{0.2812} \\
  \cline{2-5}
  \multicolumn{1}{c|}{$k$-NN}
  & \multicolumn{1}{c|}{\text{ML-GPI}} & 0.4548$\pm$0.0738 & 0.5758 & \multicolumn{1}{c}{0.3636} \\
                           \hline
\end{tabular}
\label{T:AD}
\end{table}


\section{Conclusions and discussion}

In this paper, we proposed a novel nonlinear metric learning model
through piecewise linearization. Unlike existing models, our solution
merges linear component metrics based on the velocities instead of
displacements. The setup ensures the resulted transformation to be
diffeomorphic, which enables a smooth transition crossing the
boundaries among classes.
Generating inherently smooth component metrics with tensor regularization,
as well as refining the velocity combination, are the planned future work.



\bibliographystyle{IEEEtranS}
\bibliography{icpr18}

\end{document}